\documentclass{article}

\usepackage{arxiv}

\usepackage[utf8]{inputenc} 
\usepackage[T1]{fontenc}    
\usepackage{hyperref}       
\usepackage{url}            
\usepackage{booktabs}       
\usepackage{amsfonts}       
\usepackage{amsmath}        
\usepackage{nicefrac}       
\usepackage{microtype}      
\usepackage{graphicx}
\usepackage{float}
\usepackage[numbers]{natbib}

\title{ENHANCING GAME REVIEW SENTIMENT CLASSIFICATION ON STEAM PLATFORM WITH ATTENTION-BASED BILSTM}

\date{}

\author{
Abit Ahmad Oktarian \\
Department of Data Science \\
Institut Teknologi Sumatera \\
Lampung Selatan, Indonesia \\
\texttt{abit.122450042@student.itera.ac.id}
\And
Fadhil Fitra Wijaya \\
Department of Data Science \\
Institut Teknologi Sumatera \\
Lampung Selatan, Indonesia \\
\texttt{fadil.122450082@student.itera.ac.id}
\And
Dhafin Razaqa Luthfi \\
Department of Data Science \\
Institut Teknologi Sumatera \\
Lampung Selatan, Indonesia \\
\texttt{dhafin.122450133@student.itera.ac.id}
\And
Luluk Muthoharoh, M.S \\
Department of Data Science \\
Institut Teknologi Sumatera \\
Lampung Selatan, Indonesia \\
\texttt{luluk.muthoharoh@sd.itera.ac.id}
\And
Ardika Satria, M.Si \\
Department of Data Science \\
Institut Teknologi Sumatera \\
Lampung Selatan, Indonesia \\
\texttt{ardika.satria@sd.itera.ac.id}
\And
Martin Clinton Tosima Manullang, Ph.D. \\
Department of Informatics Engineering \\
Institut Teknologi Sumatera \\
Lampung Selatan, Indonesia \\
\texttt{martin.manullang@if.itera.ac.id}
}

\renewcommand{\shorttitle}{Steam Review Sentiment Classification with Attention-Based BiLSTM}

\hypersetup{
pdftitle={ENHANCING GAME REVIEW SENTIMENT CLASSIFICATION ON STEAM PLATFORM WITH ATTENTION-BASED BILSTM},
pdfsubject={cs.CL, cs.LG},
pdfauthor={Abit Ahmad Oktarian, Fadhil Fitra Wijaya, Dhafin Razaqa Luthfi, Martin Clinton Tosima Manullang, Ph.D.},
pdfkeywords={Sentiment Analysis, Steam Reviews, Deep Learning, BiLSTM, Attention Mechanism},
}

\begin{document}
\maketitle

\begin{abstract}
The rapid growth of the online gaming industry has increased the importance of user reviews as a source of information for players and developers. On the Steam platform, millions of reviews contain opinions about gameplay quality, performance, and user experience. However, the large number of reviews and the use of informal language make manual analysis inefficient. This study aims to classify the sentiment of Steam game reviews using a Deep Learning model based on Bidirectional Long Short-Term Memory (BiLSTM) with an Attention Mechanism. A total of 50,000 review samples were used, with preprocessing steps including text cleaning, tokenization, and sequence padding. The model was trained using Adam optimizer and class-weighted loss to handle data imbalance. Experimental results show that the proposed model achieved 83\% accuracy and 85\% weighted F1-score. In addition, the model reached 90\% recall for negative reviews, showing strong performance in detecting critical feedback. The Attention mechanism also improved interpretability by identifying important sentiment-related words. These results indicate that the BiLSTM+Attention model is effective for analyzing sentiment in Steam reviews and can help developers understand user feedback efficiently.
\end{abstract}


\section{Introduction}

The online gaming industry has become one of the sectors that has experienced rapid growth over the past few decades\cite{mulachela2020analisis}. Steam also provides a user review feature that contains players' opinions regarding game quality, performance, gameplay, and overall gaming experience. This feature allows users to share their experiences after playing a game, both in the form of positive and negative evaluations. In addition, the large number of available reviews can serve as an important source of information for potential players before deciding to purchase or play a particular game. In this digital era, the role of user reviews has become increasingly significant in shaping consumer perceptions of a product or service\cite{alfawas2024penerapan}. With the growing dependence of consumers on reviews and user opinions, sentiment analysis has become a vital tool for understanding customer perceptions in real-time toward certain games\cite{haris2024analisis}. Therefore, sentiment analysis is considered one of the methods for making decisions from reviews without having to read all reviews in their entirety\cite{pangestu2023analisis}. Therefore, sentiment analysis can help companies and users understand opinion trends emerging from thousands of reviews more quickly and efficiently.

However, the main challenge in sentiment analysis lies in data preprocessing, which often involves inconsistent, messy, and redundant text, requiring significant resources and frequently consuming considerable time\cite{putri2023studi}. This condition makes the text classification process more complex, so a preprocessing stage is needed to clean the data before modeling. The preprocessing stage generally includes cleaning special characters, removing stopwords, normalizing informal words, stemming, and tokenization so that the text data becomes more structured and ready to be used in the model training process. In this project, a comparison was conducted between two approaches, namely Machine Learning using TF-IDF and PyCaret AutoML, and Deep Learning using a BiLSTM + Attention architecture based on PyTorch. This comparison was carried out to examine the advantages of each method in handling diverse and dynamic Steam review data. The main objective of this study is to determine the best model for performing Steam review sentiment classification accurately and efficiently.

\section{Methodology}
\label{sec:methodology}

This section describes the dataset construction, preprocessing pipeline, model design, and evaluation protocol used in this study. The workflow is organized to support a fair comparison between a conventional machine learning baseline and the proposed deep learning model.

\subsection{Dataset}

The dataset consists of user reviews collected from the Steam platform. The original corpus contains more than 6.4 million entries in tabular format, where each row includes a review text and a binary recommendation label \cite{datasetsteam}. To balance computational cost and dataset diversity, this study uses a subset of 50,000 reviews while preserving variation across multiple game titles, including \textit{Counter-Strike} and \textit{Call of Duty}.

\subsection{Dataset Overview}

Table~\ref{tab:dataset} summarizes the dataset configuration used in this study.

\begin{table}[H]
\centering
\caption{Dataset Characteristics}
\label{tab:dataset}
\begin{tabular}{ll}
\toprule
\textbf{Attribute} & \textbf{Value} \\
\midrule
Source & Steam Reviews (Kaggle) \\
Total Original Data & 6,417,106 reviews \\
Sample Used & 50,000 reviews \\
Data Format & CSV (text + label) \\
Classes & Binary (Positive, Negative) \\
Positive Label & 1 \\
Negative Label & -1 \\
Train (DL) & 38,744 \\
Validation (DL) & 4,843 \\
Test (DL) & 4,843 \\
Test Distribution & 776 Negative, 4067 Positive \\
\bottomrule
\end{tabular}
\end{table}

Each review is mapped into one of two classes. Label 1 denotes recommended content, whereas label -1 denotes not recommended content. Review length varies substantially, ranging from short comments to multi-sentence feedback. Informal language, mixed-language expressions, and game-specific vocabulary further increase linguistic variability and make sentiment classification more challenging \cite{aggarwal2018machine}.

\subsection{Text Preprocessing}

Before modeling, each review is normalized through a sequential preprocessing pipeline. First, all text is converted to lowercase to reduce vocabulary size. Next, URL and mention patterns are removed using regular expressions. Non-alphabetic characters are then discarded, preserving only characters in the range a--z, followed by whitespace normalization. These steps reduce noise in user-generated content and produce more consistent inputs for both sparse feature extraction and neural sequence modeling.

\subsection{Machine Learning Baseline}

As a baseline approach, the reviews are represented using \textit{term frequency--inverse document frequency} (TF-IDF). Under this representation, each document is converted into a sparse vector whose weights reflect both term frequency within a document and inverse document frequency across the corpus \cite{salton1988term}.

Model selection is conducted using PyCaret under a uniform experimental configuration. Stratified 3-fold cross-validation is applied to compare five classifiers: Logistic Regression, Ridge Classifier, Gradient Boosting, AdaBoost, and LightGBM. Among these candidates, LightGBM achieves the best F1-score of 0.9448 with an average training time of 11.47 seconds per fold. Ridge Classifier attains an F1-score of 0.9401 but requires substantially longer training time. These results are consistent with prior findings that gradient boosting methods perform well on high-dimensional sparse TF-IDF features \cite{ke2017lightgbm}. The selected baseline model is exported as a \texttt{.pkl} file for inference.

\subsection{Proposed BiLSTM-Attention Model}

\subsubsection{Word Embedding Layer}

For the deep learning pipeline, each tokenized review is converted into a sequence of token indices. Given a vocabulary size of $|V| = 20{,}000$ and a maximum sequence length of $L \leq 100$, each token index $x_i$ is mapped to a dense vector $e_i \in \mathbb{R}^{d_e}$ through an embedding matrix $E \in \mathbb{R}^{|V| \times d_e}$, where the embedding dimension is set to $d_e = 128$. A dropout rate of 0.3 is applied to the embedding output to reduce overfitting.

\subsubsection{Sequence Encoder}

The contextual representation is learned using a bidirectional long short-term memory (BiLSTM) network implemented in PyTorch. A forward LSTM processes the sequence from left to right, while a backward LSTM processes it in reverse order. The hidden state at time step $t$ is defined as the concatenation of the two directional outputs \cite{hochreiter1997lstm}:

\begin{equation}
h_t = [\overrightarrow{h_t}; \overleftarrow{h_t}]
\end{equation}

The hidden dimension is set to 256, resulting in 4,929,027 trainable parameters in the full model.

\subsubsection{Attention Mechanism}

Because not all tokens contribute equally to review polarity, an attention mechanism is introduced to emphasize sentiment-bearing words. For each hidden state $h_t$, an intermediate attention representation is computed as

\begin{equation}
    u_t = \tanh(W_w h_t + b_w)
\end{equation}

and the normalized attention weight is obtained using

\begin{equation}
    \alpha_t = \frac{\exp(u_t)}{\sum_{j=1}^{L} \exp(u_j)}.
\end{equation}

Here, $W_w$ and $b_w$ denote the trainable attention parameters. The final sentence representation is then calculated as a weighted sum of all hidden states:

\begin{equation}
    s = \sum_{t=1}^{L} \alpha_t h_t.
\end{equation}

This context vector emphasizes the most informative tokens in the review and improves interpretability during analysis.

\subsubsection{Classification Layer}

The context vector $s$ is passed through a dropout layer and a fully connected layer to produce logits for two sentiment classes, namely negative and positive. The resulting scores are converted into class probabilities through the softmax function.

\subsection{Training Configuration}

The deep learning model is trained using the Adam optimizer with a learning rate of $1 \times 10^{-3}$ and a batch size of 64. Because the dataset is imbalanced (42,036 positive and 7,834 negative reviews), class-weighted cross-entropy loss is employed to penalize minority-class errors more strongly. Early stopping with a patience of 3 epochs is applied based on validation loss to limit overfitting.

For reproducibility, the main hyperparameters are summarized as follows: vocabulary size $20{,}000$, maximum sequence length $100$, embedding dimension $128$, hidden dimension $256$, dropout rate $0.3$, batch size $64$, and Adam optimization with learning rate $1 \times 10^{-3}$.

\subsection{Evaluation Metrics}

Model performance is measured using Accuracy, Precision, Recall, and F1-score. These metrics are widely used in binary classification problems and are particularly informative under class imbalance \cite{powers2011evaluation}. The same evaluation protocol is used for both the machine learning baseline and the proposed BiLSTM-Attention model to maintain consistency across experiments.

\section{Training and Evaluation Framework}
\label{sec:training_eval}

\subsection{Sentiment Classification Results}

\begin{figure}[htbp]
    \centering
    \includegraphics[width=0.8\linewidth]{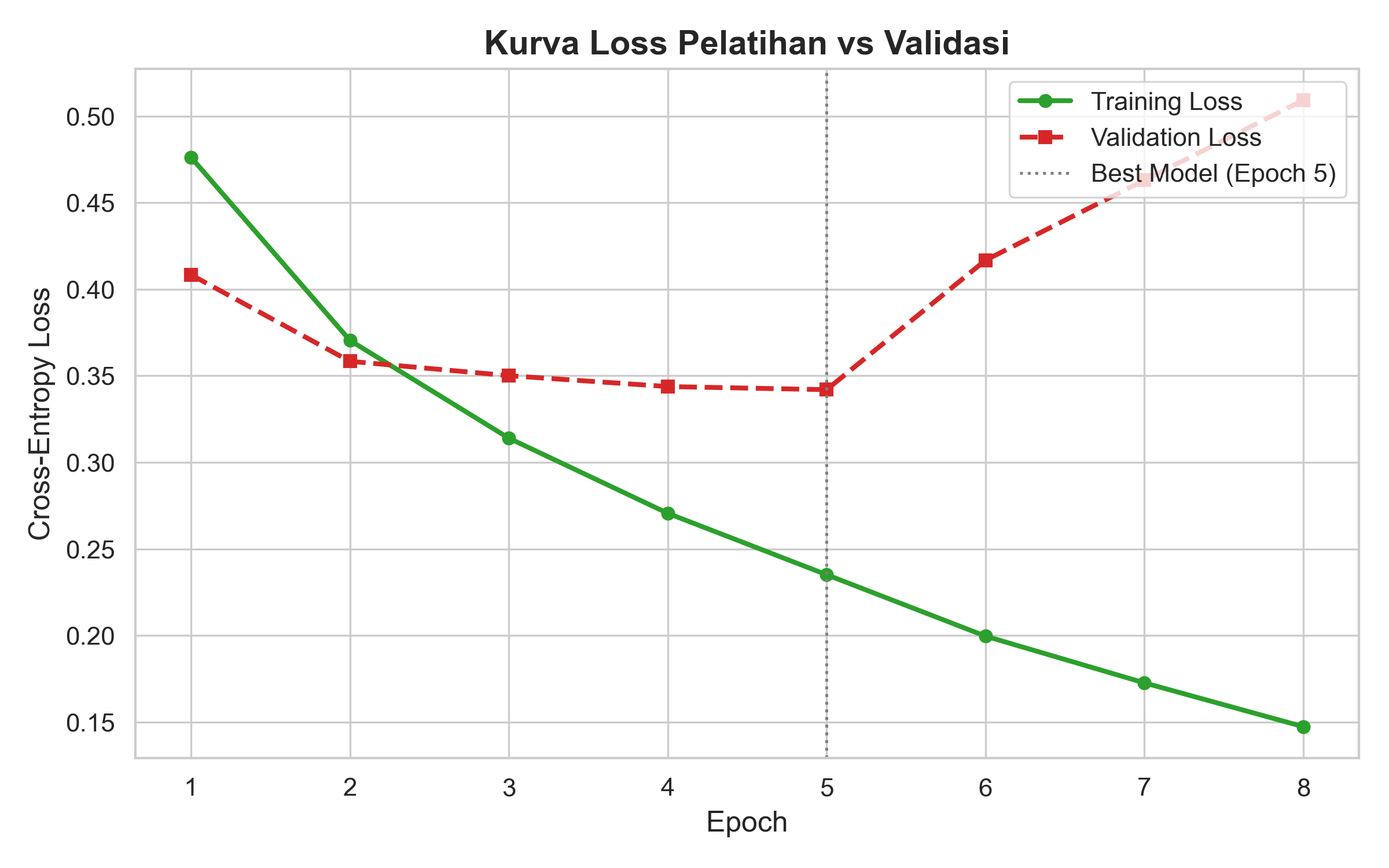}
    \caption{Training and validation loss curves over epochs.}
    \label{fig:loss_curve}
\end{figure}

As illustrated in Figure~\ref{fig:loss_curve}, the progression of the training and validation loss over eight epochs provides insight into the learning dynamics of the proposed model. The training loss declines consistently from approximately 0.48 to nearly 0.15, indicating that the network effectively minimizes prediction error on the training set.

The validation loss decreases until Epoch 5 and then rises in later epochs, indicating overfitting beyond that point. Therefore, early stopping preserves the model checkpoint obtained at Epoch 5, which provides the best validation performance.

\begin{figure}[htbp]
    \centering
    \includegraphics[width=0.6\linewidth]{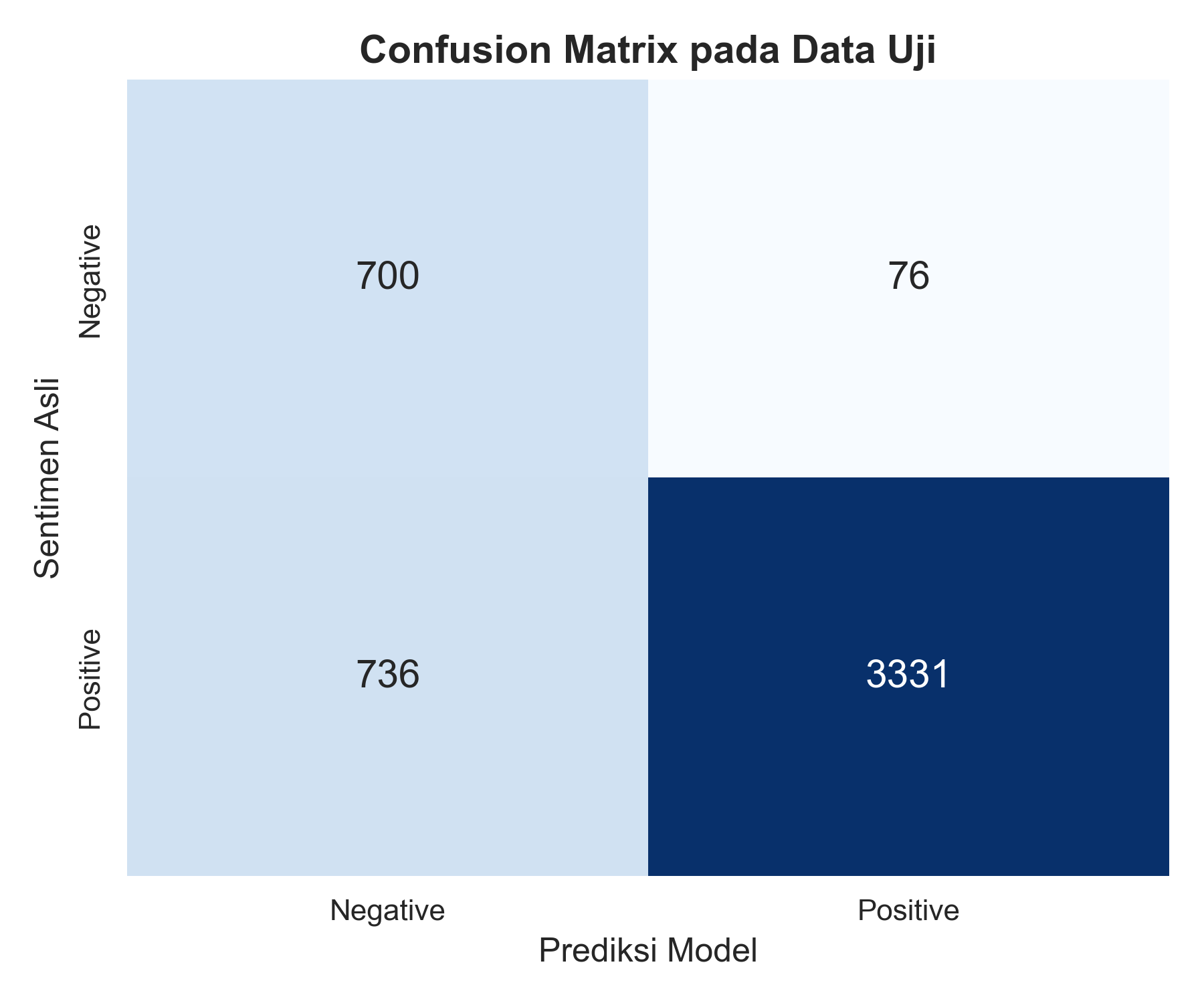}
    \caption{Confusion matrix of the BiLSTM+Attention model on the test set.}
    \label{fig:cm}
\end{figure}

The confusion matrix shown in Figure~\ref{fig:cm} further demonstrates the predictive behavior of the model. Despite the imbalanced class distribution, the use of class weights enables the model to achieve a recall of 0.90 for the negative class while maintaining an overall weighted F1-score of 0.85.

\subsection{Attention Visualization Analysis}

One advantage of the attention mechanism is improved interpretability. By inspecting the attention weights $\alpha_t$, it is possible to identify which words the model emphasizes when predicting sentiment.

\begin{figure}[htbp]
    \centering
    \includegraphics[width=0.9\linewidth]{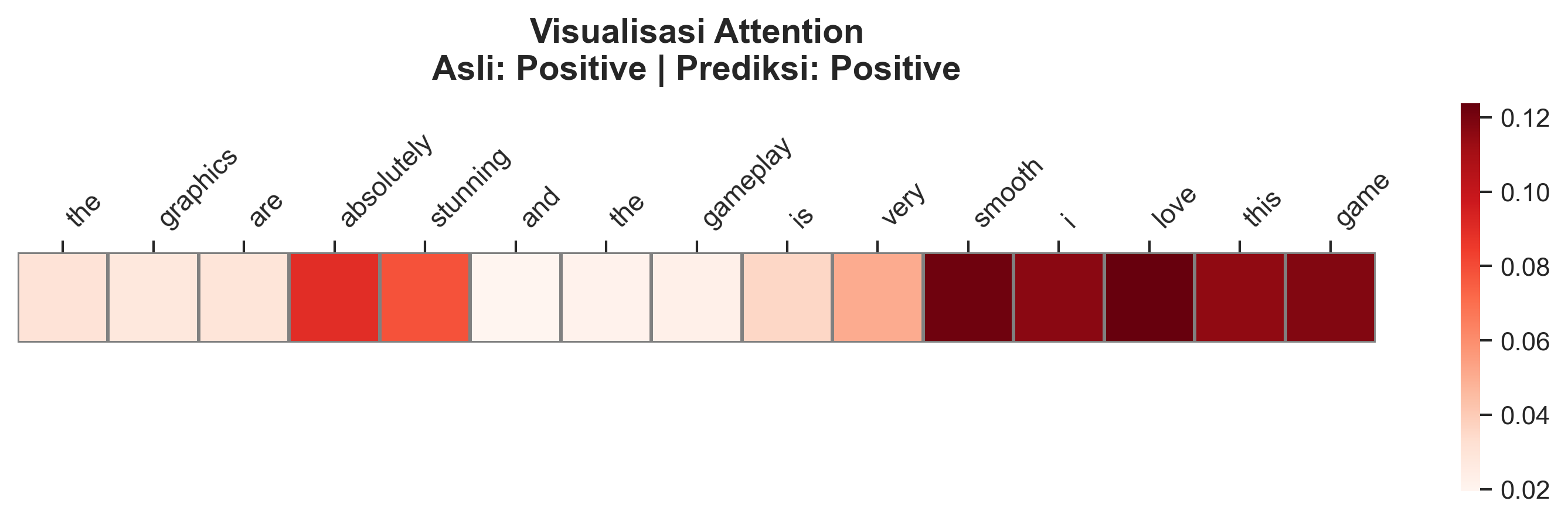}
    \vspace{0.2cm}
    \includegraphics[width=0.9\linewidth]{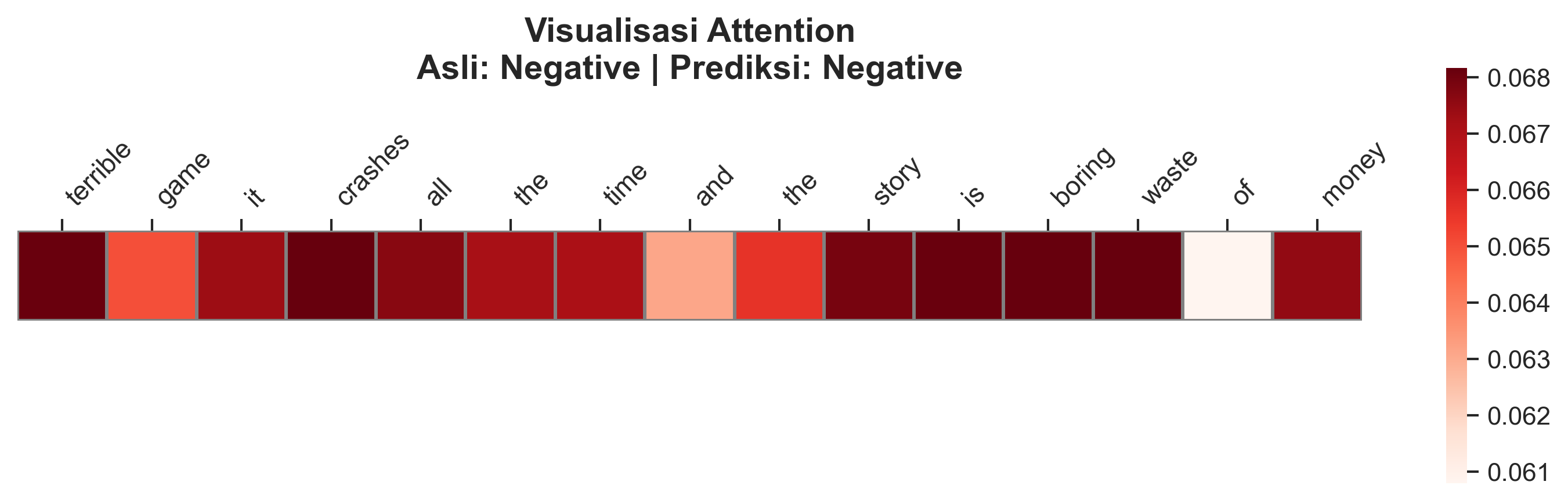}
    \caption{Attention heatmaps highlighting the model's focus on sentiment-bearing words for positive (top) and negative (bottom) reviews.}
    \label{fig:attention}
\end{figure}

Figure~\ref{fig:attention} shows that the model assigns higher attention weights to sentiment-bearing tokens. In the positive example, words such as \textit{stunning}, \textit{smooth}, and \textit{love} receive stronger emphasis. In the negative example, the model focuses on words such as \textit{terrible}, \textit{crashes}, and \textit{boring}. This behavior suggests that the BiLSTM-Attention architecture captures lexicon-level cues that are relevant to review polarity.

\subsection{Detailed Classification Metrics}

To provide a more complete view of model performance, Table~\ref{tab:classification_report} presents the classification report, including precision, recall, and F1-score for each class, along with macro and weighted averages.

\begin{table}[htbp]
    \caption{Classification Report for BiLSTM+Attention Model}
    \centering
    \begin{tabular}{lcccc}
        \toprule
        \textbf{Class} & \textbf{Precision} & \textbf{Recall} & \textbf{F1-Score} & \textbf{Support} \\
        \midrule
        Negative & 0.49 & 0.90 & 0.63 & 776 \\
        Positive & 0.98 & 0.82 & 0.89 & 4067 \\
        \midrule
        \textbf{Accuracy} & \multicolumn{3}{c}{--} & 0.83 (4843) \\
        \textbf{Macro Avg} & 0.73 & 0.86 & 0.76 & 4843 \\
        \textbf{Weighted Avg} & 0.90 & 0.83 & 0.85 & 4843 \\
        \bottomrule
    \end{tabular}
    \label{tab:classification_report}
\end{table}

\section{Conclusion}
\label{sec:conclusion}
This study presented a Deep Learning approach for classifying sentiment in Steam game reviews using a Bidirectional LSTM network integrated with an Attention Mechanism. The experimental results demonstrate that the proposed architecture effectively handles the informal language and domain-specific vocabulary prevalent in gamer communities. By utilizing class-weighted loss, the model achieved a high recall of 90\% for negative reviews, making it a highly practical tool for developers to capture critical player feedback. Furthermore, the visualization of attention weights confirmed the model's interpretability, proving its ability to accurately isolate sentiment-bearing keywords. Future work could explore the integration of transformer-based architectures or multilingual embeddings to process non-English reviews on the platform.

\section*{Acknowledgments}
This project was completed as a final assignment for the Natural Language Processing (Pengolahan Bahasa Alami) course at Institut Teknologi Sumatera (ITERA). The authors extend their gratitude to the course instructors for their guidance and support throughout the development of this research.

\bibliographystyle{IEEEtran}
\bibliography{references}  

@article{mulachela2020analisis,
  title={Analisis Perkembangan Industri Game di Indonesia Melalui Pendekatan Rantai Nilai Global (Global Value Chain)},
  author={Mulachela, Abdurrahman and Rizki, Khairur and Wahyuddin, YA},
  journal={Indonesian Journal of Global Discourse},
  volume={2},
  number={2},
  pages={32--51},
  year={2020}
}

@article{alfawas2024penerapan,
  title={Penerapan Fitur Ekstraksi TF-IDF untuk Analisis Sentimen Ulasan Game Bus Simulator Indonesia dengan Algoritma Naive Bayes},
  author={Alfawas, Thoriq Ikhwan and Rahim, Abdul and Rudiman, Rudiman},
  journal={Innovative: Journal Of Social Science Research},
  volume={4},
  number={5},
  pages={3177--3193},
  year={2024}
}

@article{haris2024analisis,
  title={Analisis sentimen pada game eFootball di Google Play Store menggunakan algoritma IndoBERT},
  author={Haris, Muhammad and Suharso, Aries and Nurkifli, E Haodudin},
  journal={JATI (Jurnal Mahasiswa Teknik Informatika)},
  volume={8},
  number={6},
  pages={12108--12121},
  year={2024}
}

@article{pangestu2023analisis,
  title={Analisis Sentimen Review Publik Pengguna Game Online Pada Platform Steam Menggunakan Algoritma Na{\"\i}ve Bayes},
  author={Pangestu, Adhi and Arifin, Yoseph Tajul and Safitri, Rizky Ade},
  journal={JATI (Jurnal Mahasiswa Teknik Informatika)},
  volume={7},
  number={6},
  pages={3106--3113},
  year={2023}
}

@article{putri2023studi,
  title={Studi Empiris Model BERT dan DistilBERT Analisis Sentimen pada Pemilihan Presiden Indonesia},
  author={Putri, Mahira and Sutanto, Taufik Edy and Inna, Suma},
  journal={The Indonesian Journal of Computer Science},
  volume={12},
  number={5},
  year={2023}
}

@book{aggarwal2018machine,
  title={Machine Learning for Text},
  author={Aggarwal, Charu C.},
  year={2018},
  publisher={Springer}
}

@article{salton1988term,
  title={Term-weighting approaches in automatic text retrieval},
  author={Salton, Gerard and Buckley, Christopher},
  journal={Information Processing \& Management},
  volume={24},
  number={5},
  pages={513--523},
  year={1988},
  publisher={Elsevier}
}

@inproceedings{ke2017lightgbm,
  title={LightGBM: A Highly Efficient Gradient Boosting Decision Tree},
  author={Ke, Guolin and Meng, Qi and Finley, Thomas and Wang, Taifeng and Chen, Wei and Ma, Weidong and Ye, Qiwei and Liu, Tie-Yan},
  booktitle={Advances in Neural Information Processing Systems 30 (NIPS 2017)},
  pages={3146--3154},
  year={2017}
}

@article{hochreiter1997lstm,
  title={Long Short-Term Memory},
  author={Hochreiter, Sepp and Schmidhuber, Jürgen},
  journal={Neural Computation},
  year={1997}
}

@article{powers2011evaluation,
  title={Evaluation: From Precision, Recall and F-Measure to ROC},
  author={Powers, David},
 journal={Journal of Machine Learning Technologies},
  year={2011}
}

@misc{datasetsteam,
  author={Pham, Quoc-Anh and others},
  title={Steam Reviews Dataset},
  year={2021},
  howpublished={Kaggle dataset},
  note={Contains 6,417,106 Steam reviews with recommendation labels}
}

\end{document}